\documentclass[letterpaper, 10 pt, conference]{ieeeconf}  

\IEEEoverridecommandlockouts                              

\usepackage{graphics} 
\usepackage{epsfig} 
\usepackage{times} 
\usepackage{amsmath} 
\usepackage{amssymb}  
\usepackage{colortbl}
\usepackage{xcolor}
\usepackage{multirow} 
\usepackage{booktabs}
\usepackage{arydshln} 
\usepackage[colorlinks=true, linkcolor=black, citecolor=black, urlcolor=black]{hyperref}

\title{\LARGE \bf
RGB-Only Gaussian Splatting SLAM for Unbounded Outdoor Scenes
}

\author{Sicheng Yu$^{1*}$, Chong Cheng$^{1*}$, Yifan Zhou$^1$, Xiaojun Yang$^1$, Hao Wang$^{1\dagger}$
\thanks{* Authors contributed equally to this work.}
\thanks{$^{1}$ The Hong Kong University of Science and Technology (GuangZhou).}
\thanks{$\dagger$ Corresponding author. {\tt\small haowang@hkust-gz.edu.cn}}
}

\begin{document}

\maketitle
\thispagestyle{empty}
\pagestyle{empty}

\begin{abstract}
3D Gaussian Splatting (3DGS) has become a popular solution in SLAM, as it can produce high-fidelity novel views. However, previous GS-based methods primarily target indoor scenes and rely on RGB-D sensors or pre-trained depth estimation models, hence underperforming in outdoor scenarios. To address this issue, we propose a RGB-only gaussian splatting SLAM method for unbounded outdoor scenes—OpenGS-SLAM. Technically, we first employ a pointmap regression network to generate consistent pointmaps between frames for pose estimation. Compared to commonly used depth maps, pointmaps include spatial relationships and scene geometry across multiple views, enabling robust camera pose estimation.
Then, we propose integrating the estimated camera poses with 3DGS rendering as an end-to-end differentiable pipeline. Our method achieves simultaneous optimization of camera poses and 3DGS scene parameters, significantly enhancing system tracking accuracy. Specifically, we also design an adaptive scale mapper for the pointmap regression network, which provides more accurate pointmap mapping to the 3DGS map representation. Our experiments on the Waymo dataset demonstrate that OpenGS-SLAM reduces tracking error to \textbf{9.8\%} of previous 3DGS methods, and achieves \textbf{state-of-the-art results} in novel view synthesis. Project page: \url{https://3dagentworld.github.io/opengs-slam/}.
\end{abstract}

\section{INTRODUCTION}
Simultaneous Localization and Mapping (SLAM) is a core task in the field of computer vision, extensively applied in autonomous driving, robotics, and virtual reality (VR) \cite{zhu2024nicer}. The issue of 3D representation has been a major focus, with the long-term goal of achieving both high-fidelity visual effects and precise localization capabilities \cite{matsuki2024gaussianmonogs, sandstrom2024splat-slam}. 

Previous works can be categorized into two branches, one is dense representation based methods \cite{6162880, Schops_2019_CVPR, inproceedings}, another is neural implicit representation based methods \cite{yen2021inerf,sucar2021imap,zhu2022niceslam,yang2022voxfusion,johari2023eslam,zhang2023goslam}. 
Although the former renders observed regions effectively, they fall short in novel view synthesis capability. The latter approach develops end-to-end differentiable dense visual SLAM systems, presenting strong performance. However, they have limitations such as low computational efficiency and lack of explicit modeling pose.

Recently, there are studies attempting to employ 3D Gaussian Splatting (3DGS) \cite{kerbl20233dgs} for scene representation \cite{matsuki2024gaussianmonogs,keetha2024splatam,huang2024photo,yan2024gs-slam}, which not only enables high-fidelity novel view synthesis but also achieves real-time rendering with lower memory requirements. 
Existing studies either rely heavily on high-quality depth inputs or only work on the scenarios of small-scale indoor scenes with limited camera movement \cite{matsuki2024gaussianmonogs,sandstrom2024splat-slam}.
Using RGB-only data for unbounded outdoor scenes remains challenging, due to: 1) difficulties in accurate depth and scale estimation, which impact pose accuracy and 3DGS initialization; 2) limited image overlap and singular viewing angles that lack effective constraints, leading to difficulties in training convergence.

To address the challenges above, this paper proposes a novel 3DGS-based SLAM method for unbounded outdoor scenes, \textbf{OpenGS-SLAM}. Our method only adopts RGB information, using 3DGS to represent the scene and generate high-fidelity images.

Specifically, we employ a pointmap regression network to generate consistent pointmaps between frames. These pointmaps store 3D structures from multiple standard views, which contain viewpoint relationships, 2D-to-3D correspondences, and scene geometry. This enables more robust camera pose estimation, effectively alleviating the inaccuracy issues of pre-trained depth networks \cite{wang2024dust3r}.

Furthermore, we integrate camera pose estimation with 3DGS rendering into an end-to-end differentiable pipeline. By this way, we achieve joint optimization of pose and 3DGS parameters, significantly enhancing system tracking accuracy. We also design an adaptive scale mapper and a dynamic learning rate adjustment strategy, which more accurately maps pointmap to the 3DGS map representation. 

Notably, our experiments on the Waymo dataset demonstrate OpenGS-SLAM reduces tracking error to \textbf{9.8\%} of the existing 3DGS method \cite{matsuki2024gaussianmonogs}. We also establish a new benchmark in novel view synthesis, achieving state-of-the-art results.

Our main contributions include:
\begin{enumerate}
    \item To the best of our knowledge, we are the first to propose a RGB-only 3D Gaussian Splatting SLAM method for unbounded outdoor scenes.
    \item We propose a system integrating a pointmap regression network with an end-to-end pipeline from pose estimation to 3DGS rendering. This allows for simultaneous optimization of pose and scene parameters, significantly improving tracking accuracy and stability.
    \item With the proposed adaptive scale mapper and dynamic learning rate adjustment, our OpenGS-SLAM achieves state-of-the-art performance in novel view synthesis on the Waymo dataset.
\end{enumerate}

\section{RELATED WORK}
\subsection{Differentiable Rendering SLAM.} 
Since NeRF \cite{yen2021inerf} is proposed, many NeRF-based SLAM methods have emerged. iMAP \cite{sucar2021imap} innovatively introduces NeRF as a scene representation in SLAM, utilizing a dual-threaded approach to perform camera pose tracking and scene mapping simultaneously. NICE-SLAM \cite{zhu2022niceslam} incorporates hierarchical scene representation to fuse multi-level local information. Vox-Fusion \cite{yang2022voxfusion} combines traditional volumetric fusion methods with neural implicit representations. ESLAM \cite{johari2023eslam} implements multi-scale axis-aligned feature planes, diverging from traditional voxel grids, significantly improving frame processing speed. Subsequent works, such as GO-SLAM \cite{zhang2023goslam}, HI-SLAM \cite{zhang2023hislam}, and Loopy-SLAM \cite{liso2024loopyslam}, incorporates loop closure and global bundle adjustment (BA) into NeRF-based SLAM.

Recently, due to the fast and differentiable rendering capabilities of 3DGS \cite{kerbl20233dgs}, as well as its strong scene representation, some works have begun exploring Gaussian-based SLAM. SplaTAM \cite{keetha2024splatam} integrates Gaussian scene representation into the SLAM process, optimizing camera pose and Gaussian map by minimizing rendering photometric and depth losses. MonoGS \cite{matsuki2024gaussianmonogs} derives the pose Jacobian matrix for tracking optimization and introduced isotropic regularization to ensure geometric consistency. Photo-SLAM \cite{huang2024photo} combines ORB-SLAM3 \cite{campos2021orb-slam3} with Gaussian scene representation. GS-SLAM \cite{yan2024gs-slam} proposes a robust coarse-to-fine camera tracking technique to improve tracking speed and accuracy. Gaussian-SLAM \cite{yugay2023gaussianslam} introduces submap-based Gaussian scene representation, while CG-SLAM \cite{hu2024cg-slam} uses a novel uncertainty-aware 3D Gaussian field for consistent and stable tracking and mapping.

\begin{figure*}
    \centering
    \includegraphics[width=0.9\linewidth]{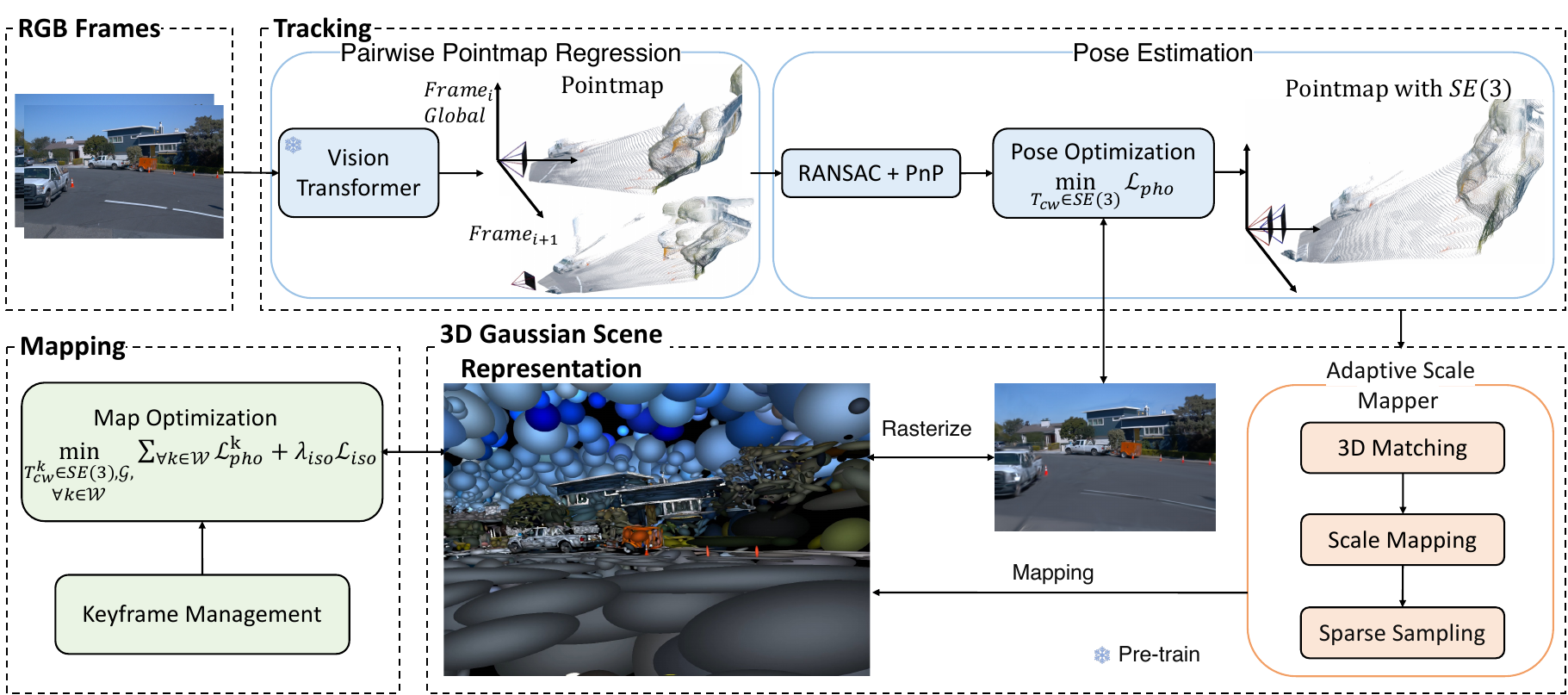}
    \caption{\textbf{SLAM System Pipeline:} Each frame inputs an RGB image for tracking. The current and previous frames are input as a pair into the Pointmap Regression network for pose estimation, followed by pose optimization based on the current Gaussian map. At keyframes, mapping is performed and the pointmap is processed by the Adaptive Scale Mapper for new Gaussian mapping. Camera pose and Gaussian map are jointly optimized in the local window.}
    \label{main}
\vspace{-15pt}
\end{figure*}

\subsection{RGB-only Dense Visual SLAM. }

Despite the success of these methods with RGB-D inputs, RGB-only SLAM presents unique challenges, primarily due to the lack of direct depth information which complicates geometric reconstruction. However, the increased difficulty makes RGB-only dense SLAM research more valuable. NeRF-SLAM \cite{rosinol2023nerfslam} and Orbeez-SLAM \cite{chung2023orbeez-slam} utilize DROID-SLAM \cite{teed2021droid-slam} and ORB-SLAM2 \cite{mur2017orb-slam2} as tracking modules, respectively, both leveraging Instant-NGP \cite{muller2022instant-ngp} for volumetric neural radiance field mapping. DIM-SLAM \cite{li2023densedim-slam} and NICER-SLAM \cite{zhu2024nicer} perform tracking and mapping on the same neural implicit map represented by hierarchical feature grids, but do not address global map consistency, such as loop closure. GO-SLAM \cite{zhang2023goslam} and Hi-SLAM \cite{zhang2023hislam} extend DROID-SLAM \cite{teed2021droid-slam} to the full SLAM setting by introducing online loop closure via factor graph optimization. GlORIE-SLAM \cite{zhang2024glorie} employs a flexible neural point cloud representation and introduces a novel DSPO layer for bundle adjustment, optimizing keyframe poses and depth.

Recently, some works have started using 3DGS to address the challenges of RGB-only SLAM. MonoGS \cite{matsuki2024gaussianmonogs} and Photo-SLAM \cite{huang2024photo} both support RGB-only inputs and achieve performance comparable to that of RGB-D inputs. MotionGS \cite{guo2024motiongs} implements tracking through feature extraction and a motion filter on each frame, using compressed 3D Gaussian representation to reduce memory usage. MGS-SLAM \cite{zhu2024mgs-slam} adopts DPVO \cite{teed2023deepdpvo} as tracking module and utilizes a pre-trained MVS network to estimate prior depth, adjusting its scale for Gaussian scene reconstruction. Splat-SLAM \cite{sandstrom2024splat-slam} combines GlORIE-SLAM \cite{zhang2024glorie} with Gaussian scene representation, introducing global BA into Gaussian-based SLAM. 
While substantial progress has been made, particularly in adapting SLAM technologies for indoor environments, the extension to outdoor settings remains limited. The development of robust RGB-only SLAM systems that can handle the unbounded and dynamic nature of outdoor environments is an ongoing area of research, with potential breakthroughs likely to have a significant impact on the field.

\section{METHOD}

\subsection{SLAM System Overview}
Fig. \ref{main} provides an overview of our system. In this section, we introduce our system from the following aspects: Tracking, 3DGS scene representation, and Mapping. Our proposed method is specifically designed to address the challenges of unbounded outdoor scenes, enhancing the tracking accuracy and robustness of scene reconstruction.

\subsection{Tracking}

\subsubsection{Pairwise Pointmap Regression and Pose Estimation}
\label{pointmap}
Previous works \cite{matsuki2024gaussianmonogs,zhu2022niceslam} involving 3DGS and NeRF primarily focus on indoor and small-scale scenes where camera movements are minimal and viewing angles are dense. In this scenario, NeRF or 3DGS can be directly used to regress camera poses. 
However, outdoor scenes typically involve vehicle-based photography, characterized by significant movement amplitudes and relatively sparse viewing angles. This makes direct regression of camera poses exceedingly challenging. 

Given that pointmaps contain the viewpoint relationships, 2D-to-3D correspondences, and scene geometry \cite{wang2024dust3r,wiles2020synsinendtoendviewsynthesis,wang2018mvpnetmultiviewpointregression}, we propose a novel pose estimation method based on a pairwise pointmap regression network, aimed at robust and rapid camera pose estimation for the current frame. 

The specific methodology is as follows: Assuming two input 2D images \(I^1, I^2 \in \mathbb{R}^{W\times H}\), we define the pointmap as the 3D points \(X^1, X^2 \in \mathbb{R}^{W\times H\times 3}\) corresponding to each pixel of these images. We utilize a pre-trained pointmap regression network that combines a ViT encoder, transformer decoders with self-attention and cross-attention layers, and an MLP regression head to generate pointmaps for consecutive frame images. Crucially, the sharing of information between the two image branches facilitates the correct alignment of pointmaps. The network is trained by minimizing the Euclidean distance between the predicted pointmaps and the actual points:
\begin{equation}
    L_{\text{reg}} = \sum_{v=(1,2)}\sum_{i\in D}\left\lVert \frac{1}{z}X_i^{v} - \frac{1}{\bar{z}}\bar{X}_i^{v} \right\rVert,
\end{equation}
where \(D\subseteq \{1 \ldots W\} \times \{1 \ldots H\}\) and \(z\) is a scale normalization factor, calculated as \(z = \frac{1}{2|D|}\sum_{v=(1,2)}\sum_{i\in D}\|X_i^v\|\).

Although the application of pointmaps might seem counterintuitive, it enables effective representation of 3D shapes in image space and allows for triangulation between rays from different viewpoints without being limited by the quality of depth estimation \cite{wang2024dust3r}. Subsequently, we employ the robust and well-established RANSAC \cite{RANSAC} with PnP \cite{lepetit2009epnp} to infer the relative pose \(T^k_{trans}\) between the two frames. Using this method, we calculate the pose for the \(k\)-th frame as \(T^k = T^k_{\text{trans}}T^{k-1}\).

\subsubsection{Pose Optimization}
To achieve precise camera pose tracking, we aim to establish a system where the photometric loss is differentiable with respect to the pose, calculated from the rendered image generated using the estimated pose. The camera poses $T$ is described as rotations and translations in 3D space and is represented by the special Euclidean group $SE(3)$, which is a manifold with nonlinear group structure. Since the Lie group $SE(3)$ is not closed under addition \cite{barfoot2024stateliegroup}, it complicates the use of gradient-based methods for optimization. To address this, we linearize $SE(3)$ into its corresponding Lie algebra $\mathfrak{se}(3)$, allowing the use of standard gradient descent techniques for optimization.

This linearization is achieved via the exponential mapping $\exp(\xi)$, where $\xi = (\omega, \nu)$ represents the infinitesimal generators of rotation and translation in the Lie algebra. The Jacobian matrix derived from the Lie algebra allows us to ensure the differentiability of the camera pose \(T_{CW}\) in the photometric loss function \(L_{pho}\), and to eliminate redundant computations \cite{kerbl20233dgs}. Using the chain rule, we first compute the derivatives of the 2D Gaussians \(\mathcal{N}(\mu_I, \Sigma_I)\) with respect to the camera pose \(T_{CW}\). These 2D Gaussians are obtained by applying EWA splatting \cite{zwicker2001ewa} to the 3D Gaussians \(\mathcal{N}(\mu_W, \Sigma_W)\). The derivatives of \(\mu_I\) and \(\Sigma_I\) are given by \cite{matsuki2024gaussianmonogs}.

\begin{equation}
    \frac{\partial \mu_I}{\partial T_{\text{CW}}} = \frac{\partial \mu_I}{\partial \mu_C} \cdot  [I\  -\mu_c^\times],
\end{equation}
\begin{equation}
    \frac{\partial \Sigma_I}{\partial T_{\text{CW}}} = \frac{\partial \Sigma_I}{\partial J} \cdot \frac{\partial J}{\partial \mu_C} \cdot [I\  -\mu_c^\times] + \frac{\partial \Sigma_I}{\partial W} \cdot \begin{bmatrix}
    0 & -W^\times_{:,1} \\
    0 & -W^\times_{:,2} \\
    0 & -W^\times_{:,3}
    \end{bmatrix},
\end{equation}
where ${}^\times$ denotes the skew symmetric matrix of a 3D vector, $W$ is the rotational component of $T_{CW}$, and $W^\times_{:,i}$ refers to the $i$th column of the matrix.

This calculation is essential for the differentiability of camera pose. We define the photometric loss $L_{pho}$ as:
\begin{equation}
    L_{\text{pho}} = \left\lVert r(\mathcal{G}, T_{CW}) - \bar{I} \right\rVert_{1}.
\end{equation}
where $r$ denotes per-pixel differentiable rendering function, producing the image through Gaussians $\mathcal{G}$ and camera pose $T_{CW}$, and $\bar{I}$ represents the ground truth image.

Finally, the derivative of the photometric loss $L_{\text{pho}}$ with respect to the pose $T_{CW}$ can be computed using the chain rule:
\begin{equation}
    \nabla_T L_{\text{pho}} = \frac{\partial L_{\text{pho}}}{\partial r}  \cdot \left( \frac{\partial r}{\partial \mu_I} \cdot \frac{\partial \mu_I}{\partial T_{CW}} + \frac{\partial r}{\partial \Sigma_I} \cdot \frac{\partial \Sigma_I}{\partial T_{CW}} \right)
\end{equation}

By following these steps, we link incremental pose updates through the differential of the rendering function to the photometric loss. This enables end-to-end optimization of the camera pose based on 3DGS rendering results, ensuring both high precision and robust pose tracking.

\subsection{3D Gaussian Scene Representation}
\subsubsection{3D Gaussian Map}
To achieve real-time rendering while preserving the advantages of volumetric scene representations, our SLAM system employs a 3DGS-based scene representation method \cite{kerbl20233dgs}. This technique not only enhances rendering speeds but also maintains high flexibility and precision.

In this representation, the scene is modeled using a set of Gaussians centered at the point \(\mu\), with the shape and orientation of each Gaussian described by its covariance matrix \(\Sigma\). The distribution of each Gaussian is defined as \cite{kerbl20233dgs}:
\begin{equation}
    G(x) = e^{-\frac{1}{2} (x - \mu)^T \Sigma^{-1} (x - \mu)},
\end{equation}
where \(x\) represents an arbitrary position within the 3D scene.

The covariance matrix is ingeniously decomposed into a scaling matrix \(S\) and a rotation matrix \(R\), enhancing control over the scene's geometry \cite{kerbl20233dgs}:
\begin{equation}
    \Sigma = RSS^T R^T.
\end{equation}
We omit spherical harmonics representing view-dependent radiance. By projecting the 3D Gaussians onto a 2D plane and using tile-based rasterization techniques for efficient sorting and blending, we achieve rapid and precise color rendering. The color of a pixel \(x'\) is determined by \cite{kerbl20233dgs}:
\begin{equation}
    C(x') = \sum_{i \in N} c_i \alpha_i \prod_{j=1}^{i-1} (1 - \alpha_j),
\end{equation}
where \(N\) is the set of Gaussians influencing pixel \(x'\).

Through this method, our SLAM system can swiftly adapt to dynamic environments, optimizing all Gaussian parameters, including position, rotation, scale, opacity, and color. The entire process is differentiable, facilitating straightforward adjustments and improvements.

\subsubsection{Adaptive Scale Mapper} 
We utilize the pointmap previously obtained, as referenced in \ref{pointmap}, to map the 3D Gaussian map. While the pointmap maintains consistency through shared information across branches, we propose an adaptive scale mapper to enhance robustness against inaccuracies in pointmap regression. This mapper is designed to perform scale-mapping on the pointmaps before inserting new Gaussians based on the pointmaps by performing 3D matching across consecutive frames. Specifically, we match pointmaps \(\{X^{k-1}, X^k\}\) and \(\{X'^{k}, X'^{k+1}\}\) generated from cross-view images of three frames (\(k-1\), \(k\), \(k+1\)), allowing us to measure the relative distances of the same points across frames. To quantify the scale changes between consecutive frames, we define the scale ratio \(\rho_{ij}\) as:
\begin{equation}
    \rho_{ij} = \frac{\|X_i'^{k} - X_j'^{k+1}\|}{\|X_i^{k-1} - X_j^{k}\|}
\end{equation}

where \(X_i^{k-1}\) and \(X_j^{k}\) are corresponding points in frame \(k\), and \(X_i'^{k}\) and \(X_j'^{k+1}\) are their counterparts in frame \(k+1\). This ratio reflects the scale change from frame \(k\) to \(k+1\).

By averaging multiple \(\rho_{ij}\) values, we estimate the overall scene's mean scale change. To maintain scale consistency throughout the sequence, we use the first frame as a reference and calculate the scale factor for each frame relative to the first by cumulatively multiplying the average scale ratios:

\begin{equation}
    S_k = S_{k-1} \cdot \frac{1}{N} \sum_{ij} \frac{\|X_i'^{k} - X_j'^{k+1}\|}{\|X_i^{k-1} - X_j^{k}\|}
\end{equation}

This approach ensures frame-to-frame scale consistency, enabling the scale factors to be used to map subsequent frame pointmap coordinates, crucial for precise 3D mapping and location tracking in outdoor scenes. Finally, based on our observations, not all 3D Gaussian points contribute to mapping; therefore, we introduce a sparse subsampling method. This method employs a hierarchical structure to effectively control the number of 3D Gaussian points, ensuring high-quality mapping while reducing processing time.

\subsection{Mapping}
\subsubsection{Keyframe Manegement} A good keyframe selection strategy should ensure sufficient viewpoint overlap while avoiding redundant keyframes. Since it is infeasible to jointly optimize the Gaussian scene and camera pose with all keyframes, we manage a local keyframe window $\mathcal{W}$ to select non-redundant keyframes observing the same area, providing better multi-view constraints for subsequent mapping optimization. With this in mind, we adopt the keyframe management strategy from \cite{matsuki2024gaussianmonogs}: selecting keyframes based on covisibility and managing the local window by evaluating the overlap with the most recent keyframe.
\subsubsection{Gaussian Map Optimization}
At each keyframe, we optimize the Gaussian map by jointly optimizing the Gaussian attributes and camera poses within the currently managed local keyframe window $\mathcal{W}$, performing local window BA. The optimization is still carried out by minimizing photometric loss. To reduce excessive stretching of the ellipsoids, we employed isotropic regularization \cite{matsuki2024gaussianmonogs}:
\begin{equation}
    L_{\text{iso}} = \sum_{i=1}^{|\mathcal{G}|} \left\lVert s_i - \tilde{s}_i \cdot \mathbf{1} \right\rVert_1
\end{equation}
to penalise the scaling parameters \(s_i\). The mapping optimization task can be summarized as:
\begin{equation}
    \min_{\substack{T_{CW}^k \in SE(3) \\ \forall k \in \mathcal{W}}, \mathcal{G}} \quad \sum_{\forall k \in \mathcal{W}} L_{\text{pho}}^k + \lambda_{\text{iso}} L_{\text{iso}}.
\end{equation}

\subsubsection{Adaptive Learning Rate Adjustment.} 
In classical indoor SLAM datasets, the camera typically captures a small scene in a loop, causing the learning rate for Gaussian optimization to gradually decay as the number of cumulative iterations increases. However, the outdoor data we are studying is captured by a front-facing vehicle camera in street scenes, where the traversed areas are not revisited. Therefore, a different learning rate decay strategy is required. We aim for the learning rate to gradually decay when the vehicle is traveling along straight roads, and to increase when the vehicle encounters slope or makes turns in order to optimize new scenes. To address this, we propose an adaptive learning rate adjustment based on rotation angle. We still adjust the learning rate based on cumulative iterations \(N_{iter}\), and then adaptively adjust the cumulative iterations. Suppose the current keyframe and the last keyframe have rotation matrices \(R_{1}, R_{0}\) respectively, the relative rotation matrix is \(R_{diff}=R_{0}^T R_{1}\), from which the rotation radian can be computed:
\begin{equation}
    \theta_{rad} = \cos^{-1}\left(\frac{\text{trace}(\mathbf{R}_{\text{diff}}) - 1}{2}\right). 
\end{equation}
We then convert \(\theta_{rad}\) to degrees \(\theta\) . If \(\theta>2\), we adjust the cumulative iterations:
\begin{equation}
    N_{iter}^{new} = N_{iter} \times \left( 1 - \sqrt{\frac{\theta}{90}} \right). 
\end{equation}
When the rotation reaches 90 degrees, iterations are reset. The square root adjustment ensures that small angle changes lead to a more significant increase in learning rate. The adaptive learning rate adjustment effectively improved the quality of mapping in the later stages, with detailed analysis provided in the ablation study section.

\begin{table}[!ht]
\centering
\fontsize{7}{8}\selectfont  
\setlength{\tabcolsep}{0.8pt} 
\renewcommand{\arraystretch}{1}

\definecolor{midred}{HTML}{FF9999}
\definecolor{midorange}{HTML}{FFCC99}
\definecolor{midyellow}{HTML}{FFFF99}

\caption{Tracking and Rendering results on 9 Waymo segments. ATE RMSE[m] for tracking, PSNR, SSIM, and LPIPS for rendering. Best results are highlighted as \textbf{\raisebox{-0.5ex}{\fcolorbox{white}{midred}{FIRST}}}, \textbf{\raisebox{-0.5ex}{\fcolorbox{white}{midorange}{SECOND}}}.}
\label{table:performance}

\begin{tabular}{>{\centering\arraybackslash}p{0.9cm}>{\centering\arraybackslash}p{0.9cm}*{5}{>{\centering\arraybackslash}p{1cm}}}
\toprule
\multirow{3}{*}{\centering\textbf{Segment}} & \multirow{3}{*}{\centering\textbf{Metric}} & \multicolumn{5}{c}{\textbf{Method}} \\ 
\cmidrule(lr){3-7}
& & \textbf{NICER-SLAM} & \textbf{GlORIE-SLAM} & \textbf{Photo-SLAM} & \textbf{MonoGS} & \textbf{Ours} \\
\midrule

\multirow{4}{*}{\centering \textbf{100613}} 
& \hfill ATE$\downarrow$ & 19.39 &\cellcolor{midred}0.302 & 14.28 & 6.953 & \cellcolor{midorange}0.324 \\ 
& \hfill PSNR$\uparrow$ & 11.46 & 18.58 & 14.29 & \cellcolor{midorange}21.89 & \cellcolor{midred}24.41 \\ 
& \hfill SSIM$\uparrow$ & 0.624 & 0.750 & 0.655 & \cellcolor{midorange}0.779 & \cellcolor{midred}0.811 \\ 
& \hfill LPIPS$\downarrow$ & 0.705 & 0.595 & 0.794 & \cellcolor{midorange}0.543 & \cellcolor{midred}0.360 \\ 
\hdashline
\multirow{4}{*}{\centering \textbf{13476}} 
& \hfill ATE$\downarrow$ & 8.18 & \cellcolor{midorange}0.569 & 25.85 & 3.366 & \cellcolor{midred}0.422 \\ 
& \hfill PSNR$\uparrow$ & 8.59 & 15.06 & 17.36 & \cellcolor{midorange}20.95 & \cellcolor{midred}22.29 \\ 
& \hfill SSIM$\uparrow$ & 0.507 & 0.533 & \cellcolor{midorange}0.726 & 0.723 & \cellcolor{midred}0.733 \\ 
& \hfill LPIPS$\downarrow$ & 0.817 & 0.737 & \cellcolor{midorange}0.663 & 0.693 & \cellcolor{midred}0.602 \\ 
\hdashline
\multirow{4}{*}{\centering \textbf{106762}} 
& \hfill ATE$\downarrow$ & 35.59 & \cellcolor{midred}0.405 & 58.32 & 18.16 & \cellcolor{midorange}0.893 \\ 
& \hfill PSNR$\uparrow$ & 10.46 & 20.60 & 18.95 & \cellcolor{midorange}22.24 & \cellcolor{midred}26.19 \\ 
& \hfill SSIM$\uparrow$ & 0.425 & 0.770 & 0.802 & \cellcolor{midorange}0.814 & \cellcolor{midred}0.851 \\ 
& \hfill LPIPS$\downarrow$ & 0.670 & \cellcolor{midorange}0.507 & 0.558 & 0.515 & \cellcolor{midred}0.326 \\ 
\hdashline
\multirow{4}{*}{\centering \textbf{132384}} 
& \hfill ATE$\downarrow$ & 25.22 & \cellcolor{midred}0.142 & 3.752 & 12.08 & \cellcolor{midorange}0.436 \\ 
& \hfill PSNR$\uparrow$ & 15.12 & 20.69 & 20.03 & \cellcolor{midorange}23.48 & \cellcolor{midred}26.98 \\ 
& \hfill SSIM$\uparrow$ & 0.782 & 0.790 & 0.839 & \cellcolor{midorange}0.856 & \cellcolor{midred}0.883 \\ 
& \hfill LPIPS$\downarrow$ & 0.536 & 0.453 & 0.510 & \cellcolor{midorange}0.427 & \cellcolor{midred}0.283 \\ 
\hdashline
\multirow{4}{*}{\centering \textbf{152706}} 
& \hfill ATE$\downarrow$ & 18.67 & \cellcolor{midorange}0.425 & 18.10 & 9.180 & \cellcolor{midred}0.309 \\ 
& \hfill PSNR$\uparrow$ & 11.55 & 17.87 & 17.92 & \cellcolor{midorange}22.52 & \cellcolor{midred}23.95 \\ 
& \hfill SSIM$\uparrow$ & 0.625 & 0.626 & 0.766 & \cellcolor{midorange}0.791 & \cellcolor{midred}0.802 \\ 
& \hfill LPIPS$\downarrow$ & 0.745 & 0.677 & 0.768 & \cellcolor{midorange}0.649 & \cellcolor{midred}0.533 \\ 
\hdashline
\multirow{4}{*}{\centering \textbf{153495}} 
& \hfill ATE$\downarrow$ & 15.42 & \cellcolor{midred}1.202 & 6.407 & 5.718 & \cellcolor{midorange}1.576 \\ 
& \hfill PSNR$\uparrow$ & 11.15 & 19.40 & 18.21 & \cellcolor{midorange}21.49 & \cellcolor{midred}23.66 \\ 
& \hfill SSIM$\uparrow$ & 0.487 & 0.726 & 0.730 & \cellcolor{midorange}0.782 & \cellcolor{midred}0.800 \\ 
& \hfill LPIPS$\downarrow$ & 0.743 & \cellcolor{midorange}0.568 & 0.746 & 0.635 & \cellcolor{midred}0.499 \\ 
\hdashline
\multirow{4}{*}{\centering \textbf{158686}} 
& \hfill ATE$\downarrow$ & 20.59 & \cellcolor{midred}0.589 & 21.99 & 8.396 & \cellcolor{midorange}1.076 \\ 
& \hfill PSNR$\uparrow$ & 12.65 & 18.93 & 16.96 & \cellcolor{midorange}21.25 & \cellcolor{midred}21.71 \\ 
& \hfill SSIM$\uparrow$ & 0.609 & 0.694 & 0.696 & \cellcolor{midred}0.734 & \cellcolor{midorange}0.731 \\ 
& \hfill LPIPS$\downarrow$ & 0.756 & \cellcolor{midorange}0.539 & 0.684 & 0.574 & \cellcolor{midred}0.468 \\ 
\hdashline
\multirow{4}{*}{\centering \textbf{163453}} 
& \hfill ATE$\downarrow$ & 22.68 & \cellcolor{midred}0.646 & 25.39 & 11.21 & \cellcolor{midorange}1.719 \\ 
& \hfill PSNR$\uparrow$ & 15.38 & 19.01 & 18.58 & \cellcolor{midorange}19.28 & \cellcolor{midred}21.00 \\ 
& \hfill SSIM$\uparrow$ & 0.690 & 0.732 & 0.739 & \cellcolor{midorange}0.743 & \cellcolor{midred}0.745 \\ 
& \hfill LPIPS$\downarrow$ & 0.748 & \cellcolor{midorange}0.525 & 0.694 & 0.642 & \cellcolor{midred}0.506 \\ 
\hdashline
\multirow{4}{*}{\centering \textbf{405841}} 
& \hfill ATE$\downarrow$ & 10.60 & \cellcolor{midred}0.546 & 5.466 & 1.703 & \cellcolor{midorange}0.800 \\ 
& \hfill PSNR$\uparrow$ & 13.66 & 19.32 & 17.31 & \cellcolor{midorange}23.14 & \cellcolor{midred}25.72 \\ 
& \hfill SSIM$\uparrow$ & 0.621 & 0.698 & 0.724 & \cellcolor{midorange}0.804 & \cellcolor{midred}0.840 \\ 
& \hfill LPIPS$\downarrow$ & 0.815 & 0.553 & 0.655 & \cellcolor{midorange}0.522 & \cellcolor{midred}0.333 \\ 
\hdashline
\multirow{4}{*}{\centering \textbf{Avg.}} 
& \hfill ATE$\downarrow$ & 19.59 & \cellcolor{midred}0.536 & 19.95 & 8.529 & \cellcolor{midorange}0.839 \\ 
& \hfill PSNR$\uparrow$ & 12.22 & 18.83 & 17.73 & \cellcolor{midorange}21.80 & \cellcolor{midred}23.99 \\ 
& \hfill SSIM$\uparrow$ & 0.622 & 0.702 & 0.741 & \cellcolor{midorange}0.780 & \cellcolor{midred}0.800 \\ 
& \hfill LPIPS$\downarrow$ & 0.726 & \cellcolor{midorange}0.572 & 0.674 & 0.577 & \cellcolor{midred}0.434 \\ 
\bottomrule
\label{performance}
\end{tabular}
\vspace{-20pt}
\end{table}

\section{EXPERIMENTS}

\subsection{Implementation and Experiment Setup}

\subsubsection{Datasets}
We evaluate the Waymo open dataset \cite{Sun_2020_CVPRwaymo}, focusing on tracking accuracy, novel view rendering performance comparison, and ablation study. Waymo dataset, collected by autonomous vehicles, contains outdoor street scenes. We use the front-facing RGB images captured by the vehicle's cameras as input.

\subsubsection{Baseline Methods}
We compare our method with four SLAM approaches that support monocular RGB-only input and novel view rendering, including NICER-SLAM \cite{zhu2024nicer}, GlORIE-SLAM \cite{zhang2024glorie}, Photo-SLAM \cite{huang2024photo}, and MonoGS \cite{matsuki2024gaussianmonogs}.

\subsubsection{Metrics}
To assess novel view rendering performance, we use PSNR, SSIM \cite{SSIM}, and LPIPS metrics, calculated on frames excluding keyframes (i.e., training frames). For tracking accuracy, we use ATE RMSE (in meters) as the evaluation metric.

\subsubsection{Implementation Details}
We run our SLAM on a single NVIDIA RTX A6000 GPU. As with MonoGS, rasterization and gradients computation for Gaussian attributes and camera pose are implemented through CUDA. The rest of the SLAM pipeline is developed using PyTorch. We employed the best-performing pre-trained pointmap regression network \cite{wang2024dust3r} in our tests. Local window size $|\mathcal{W}|=8$, isotropic regularization coefficient $\lambda_{iso}=10$.

\begin{figure*}
    \centering
    \includegraphics[width=0.93\linewidth]{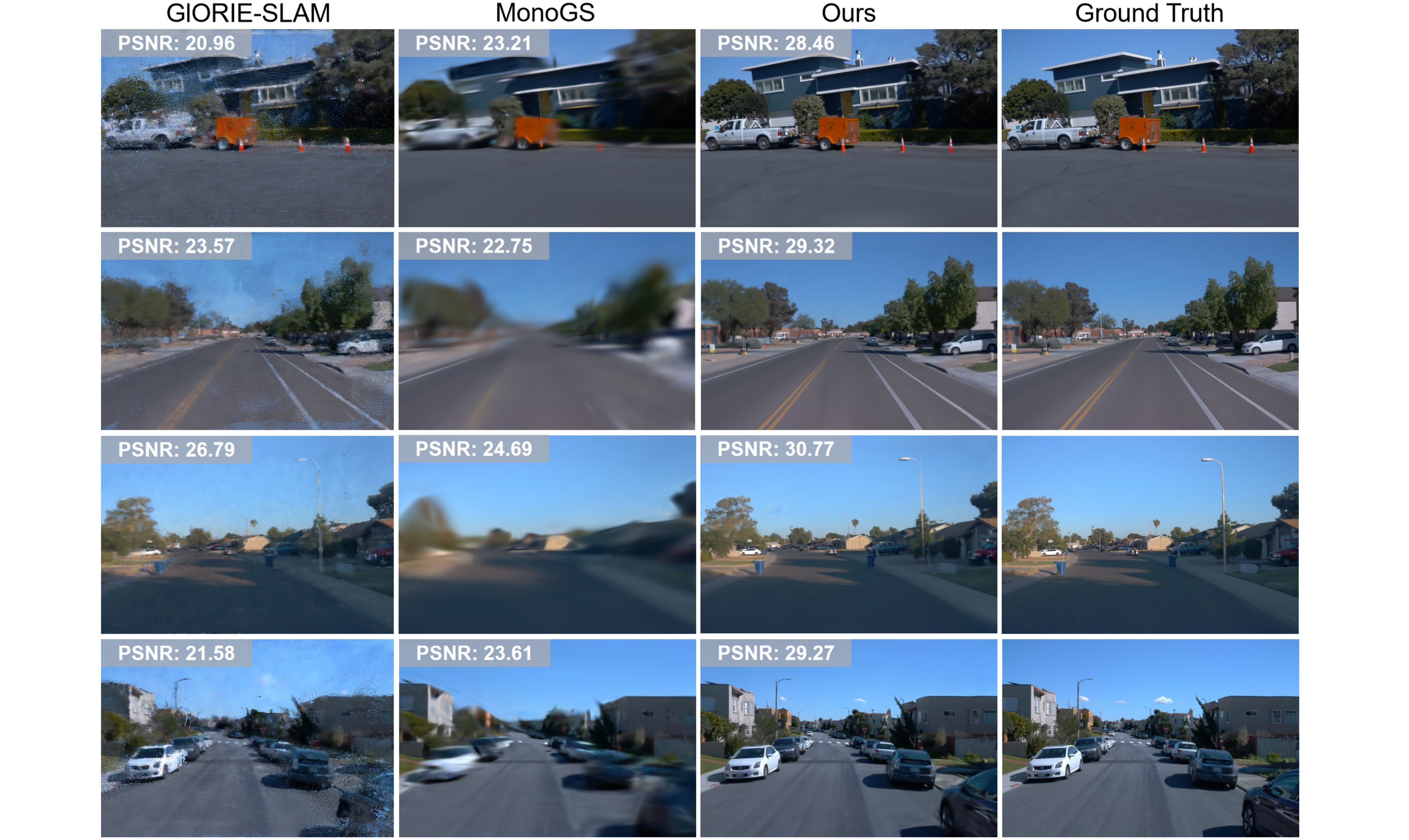}
    \caption{\textbf{Novel View Rendering Results on 4 Waymo segments.} For unbounded outdoor scenes, our method renders high-fidelity images, accurately capturing details of vehicles, streets, and buildings. In contrast, MonoGS and GlORIE-SLAM exhibit rendering distortions and blurriness.}
    \label{render1}
    \vspace{-10pt}
\end{figure*}

\subsection{Experiment Results}
\begin{figure}[t]
    \centering
    \includegraphics[width=1\linewidth]{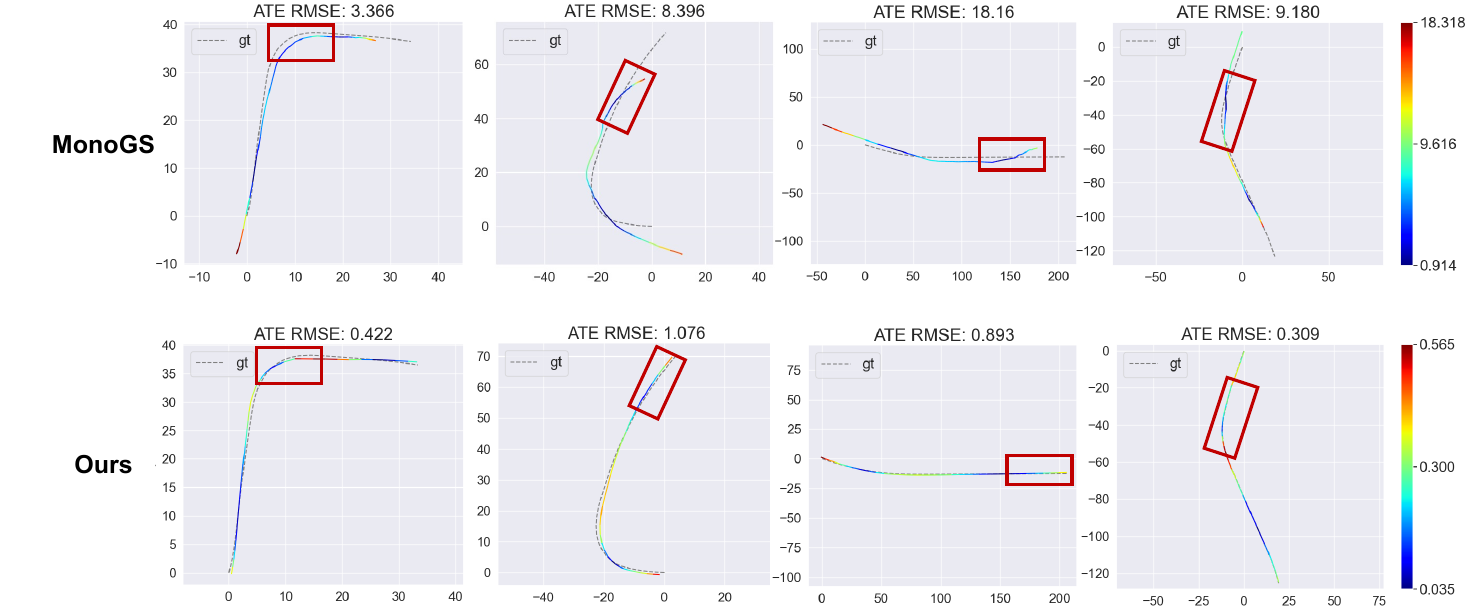}
    \caption{Comparison of tracking trajectories with MonoGS on 4 segments. Our method greatly enhances tracking accuracy, with no noticeable drift.}
    \label{monogs_tracking}
\vspace{-15pt}
\end{figure}

TABLE \ref{performance} shows the performance of tracking and novel view rendering. Our method achieves the best novel view rendering performance across all segments. Compared to MonoGS, which also use Gaussian scene representation, our approach improves the average PSNR by 10\%. In terms of tracking accuracy, our method slightly lags behind GlORIE-SLAM but significantly outperforms the other methods. GlORIE-SLAM employs a frame-to-frame tracking module based on optical flow and performs global BA every 20 frames and at the end of the system. In contrast, our tracking relies on Gaussian map-based pose alignment, which requires robust scene reconstruction—a challenging task in unbounded outdoor environments. As shown in Fig. \ref{monogs_tracking}, compared to MonoGS, which tracks in a similar manner, our tracking trajectories are noticeably more accurate, with no significant drift, and effectively handle sharp turns, demonstrating the strength of our approach. Additionally, despite not incorporating global BA or backend filtering, we achieve results comparable to GlORIE-SLAM and even outperform it on two segments, while also possessing superior NVS capabilities, highlighting the potential of our method. Moreover, implementing efficient and accurate global BA based on Gaussian scenes is no trivial task.

Fig. \ref{render1} presents the novel view rendering results, where our method renders high-fidelity images that accurately capture details of vehicles, streets, and buildings in both forward and side views. While GlORIE-SLAM achieves the best tracking performance, its rendered images suffer from missing pixels and distortions. Although MonoGS also uses Gaussian representation, its rendered images are very blurry. These results clearly demonstrate the superior novel view rendering capability of our method in unbounded outdoor scenes.

\begin{table}[h]
\centering
\setlength{\tabcolsep}{5pt} 
\caption{Ablation Study Results: Impact on tracking and novel view rendering performance after removing each module.}
\label{table:ablation}
\begin{tabular}{lcccc}
\toprule 
\textbf{Method} & \textbf{ATE RMSE } & \textbf{PSNR} & \textbf{SSIM } & \textbf{LPIPS} \\
\midrule 
w/o lr adjustment & 1.836 & 23.08 & 0.781 & 0.436 \\
w/o scale mapper  & 1.095 & 23.49 & 0.787 & 0.450 \\
w/o pointmap regression & 11.18 & 18.47 & 0.734 & 0.614 \\
\textbf{Ours}  & \textbf{0.839} & \textbf{23.99} & \textbf{0.800} & \textbf{0.434} \\
\bottomrule 
\label{ablation}
\vspace{-24pt}
\end{tabular}
\end{table}

\subsection{Ablation Study}
In this section, we demonstrate the importance of pointmap regression to the overall SLAM framework, as well as the impact of adaptive scale mapper and adaptive learning rate adjustment on performance. The average results across the 9 Waymo segments are reported in TABLE \ref{ablation}.

\subsubsection{Adapative Learning Rate Adjustment}
\begin{figure}[t!]
    \centering
    \includegraphics[width=1\linewidth]{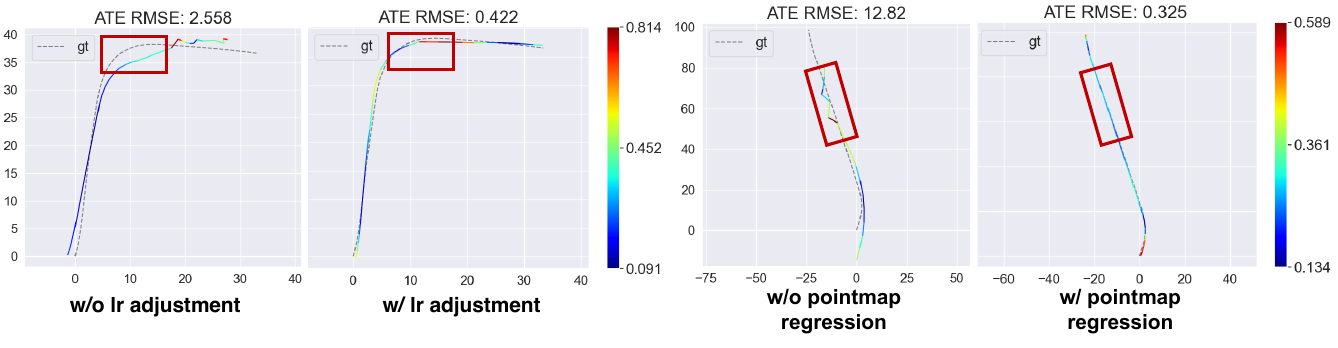}
    \vspace{-20pt}
    \caption{Ablation study of lr adjustment and pointmap regression: tracking trajectories on two segments. Without them, tracking fails during the process.}
    \label{xr1}
    \vspace{-15pt}
\end{figure}

Learning rate adjustment is crucial for tracking accuracy, particularly during turns. Fig. \ref{xr1} shows that significant trajectory drift occurs during turns without learning rate adjustment, affecting subsequent tracking. This is because after a turn, the Gaussian map for the new viewpoint requires a higher learning rate for proper adjustment, and our tracking relies on an accurate Gaussian map.

\subsubsection{Adaptive Scale Mapper}
Without adaptive scale mapper, both tracking and novel view rendering performance degrade. We have learned that pointmaps regressed from different frames have scale discrepancies, and failure to adjust these will misplace newly inserted Gaussians, negatively impacting the entire SLAM system's performance.

\subsubsection{Pairwise Pointmap Regression}
Without pointmap regression, we use the estimated pose from the previous frame as the initialization and generate depth maps through depth rasterization with added noise for Gaussian insertion. This approach produces poor results, highlighting the importance of pointmap regression, as its pre-trained information is crucial for accurate outdoor scene reconstruction.

\section{CONCLUSIONS}
In this paper, we introduce OpenGS-SLAM, an RGB-only SLAM system based on 3DGS representation for unbounded outdoor scenes. Our approach integrates a pointmap regression network with 3DGS representation, ensuring precise camera pose tracking and excellent novel view synthesis capabilities. Compared to other 3DGS-based SLAM systems, our method offers superior tracking accuracy and robustness in outdoor settings, making it highly practical for real-world applications.

\section*{Acknowledgment}
This research is supported by the National Natural Science Foundation of China (No. 62406267), Guangzhou-HKUST(GZ) Joint Funding Program (Grant No. 2025A03J3956), the Guangzhou Municipal Science and Technology Project (No. 2025A04J4070), the Guangzhou Municipal Education Project (No. 2024312122) and Education Bureau of Guangzhou Municipality.

\clearpage
\bibliographystyle{IEEEtran} 
\bibliography{references.bib}

\end{document}